\pdfoutput=1
\PassOptionsToPackage{unicode}{hyperref}
\PassOptionsToPackage{hyphens}{url}
\PassOptionsToPackage{dvipsnames,svgnames,x11names}{xcolor}
\documentclass[
]{article}
\usepackage{xcolor}
\usepackage{amsmath,amssymb}
\usepackage{iftex}
\ifPDFTeX
  \usepackage[T1]{fontenc}
  \usepackage[utf8]{inputenc}
  \usepackage{textcomp} 
\else 
  \usepackage{unicode-math} 
  \defaultfontfeatures{Scale=MatchLowercase}
  \defaultfontfeatures[\rmfamily]{Ligatures=TeX,Scale=1}
\fi
\usepackage{lmodern}
\ifPDFTeX\else
\fi
\IfFileExists{upquote.sty}{\usepackage{upquote}}{}
\IfFileExists{microtype.sty}{
  \usepackage[]{microtype}
  \UseMicrotypeSet[protrusion]{basicmath} 
}{}
\makeatletter
\@ifundefined{KOMAClassName}{
  \IfFileExists{parskip.sty}{%
    \usepackage{parskip}
  }{
    \setlength{\parindent}{0pt}
    \setlength{\parskip}{6pt plus 2pt minus 1pt}}
}{
  \KOMAoptions{parskip=half}}
\makeatother
\usepackage{longtable,booktabs,array}
\usepackage{calc} 
\usepackage{etoolbox}
\makeatletter
\patchcmd\longtable{\par}{\if@noskipsec\mbox{}\fi\par}{}{}
\makeatother
\IfFileExists{footnotehyper.sty}{\usepackage{footnotehyper}}{\usepackage{footnote}}
\makesavenoteenv{longtable}
\usepackage{graphicx}
\makeatletter
\newsavebox\pandoc@box
\newcommand*\pandocbounded[1]{
  \sbox\pandoc@box{#1}%
  \Gscale@div\@tempa{\textheight}{\dimexpr\ht\pandoc@box+\dp\pandoc@box\relax}%
  \Gscale@div\@tempb{\linewidth}{\wd\pandoc@box}%
  \ifdim\@tempb\p@<\@tempa\p@\let\@tempa\@tempb\fi
  \ifdim\@tempa\p@<\p@\scalebox{\@tempa}{\usebox\pandoc@box}%
  \else\usebox{\pandoc@box}%
  \fi%
}
\def\fps@figure{htbp}
\makeatother
\providecommand{\tightlist}{%
  \setlength{\itemsep}{0pt}\setlength{\parskip}{0pt}}
\usepackage[]{natbib}
\usepackage[preprint]{neurips_2026}
\usepackage{microtype}
\date{}
\usepackage{bookmark}
\IfFileExists{xurl.sty}{\usepackage{xurl}}{} 
\makeatletter
\@ifundefined{xmpquote}{}{}
\makeatother
\hypersetup{
  pdftitle={Trajectories That Segment Themselves: Agent-Declared Boundaries as a Training Unit},
  colorlinks=true,
  linkcolor={Maroon},
  filecolor={Maroon},
  citecolor={Blue},
  urlcolor={Blue},
  pdfcreator={LaTeX via pandoc}}

\title{Trajectories That Segment Themselves: Agent-Declared Boundaries
as a Training Unit}
\author{Jingxi Wei\\ \texttt{jingxiw3@illinois.edu}}
\date{}

\begin{document}
\maketitle
\begin{abstract}
Long-horizon coding-agent trajectories are poorly matched to the credit
units available to train on: a single action has no stable value, an
episode label merges productive exploration with abandoned directions,
and a fixed window cuts where the logging mechanics fall. We introduce
\textbf{collection-time semantic self-segmentation}, in which a
declarative contract has the acting agent expose its own boundaries
while the trajectory is generated. Instantiated with falsifiable causal
hypotheses, successive adoptions expose variable-length semantic phases,
and no milestone vocabulary, gold patch, environment replay, teacher
logits, or retrospective segmenter places a boundary. The declaration is
an interface rather than a marker: because the agent names its current
conjecture, a reviewer can negate that conjecture by name and eliminate
it, which is what lets our collection protocol manufacture
wrong-cause-then-correction transitions that recorded work rarely
contains, and one collection then yields four supervised targets ---
including audit supervision drawn from exactly the failed regions an
episode label discards. We then ask what survives deleting the
declaration. Given the cut points but not the hypothesis, a model
attributes action blocks to the hypothesis that governed them at over
twice chance --- and better than equal-length blocks over the same
trajectories, paired sign test \(p = 0.0002\) --- surviving a model-free
lexical control and collapsing under a label permutation. Asked instead
to place boundaries, a code-blind annotator matches 24 of 40 where
random placement matches 11.5, while a mechanical test-event rule beats
chance at neither end of a strict-to-permissive sweep. The segments are
therefore coherent and not cheaply reproducible. Downstream, standard
DPO fits 2,551 phase-boundary pairs: no decision changes on 91
adversarially built held-out items, while four of 60 change on
matched-construction items, all wrong to right, where two controls
change none. With 1,825 of the pairs drawn from a single generator
family, what the two sets separate is how the pairs were written, and
the variable to vary next is the diversity of the pair corpus rather
than the boundary.

\textbf{Code and data.} The pipeline is at
\url{https://github.com/Jingxi-Wei/hypothesis-ledger} and the phase
ledger with the derived training and evaluation views at
\url{https://huggingface.co/datasets/jingxiwei/hypothesis-ledger-selfcorrection}.
Appendix A states what the release leaves out and why.
\end{abstract}

\section{Introduction}\label{introduction}

\subsection{The mismatch between agent search and common training
units}\label{the-mismatch-between-agent-search-and-common-training-units}

Repository-level software tasks require agents to search through
competing causal explanations, and the value of each event in that
search depends on the causal attempt it belongs to. This leaves the
available credit units all wrong in different ways. A \textbf{single
action} such as \texttt{grep} or \texttt{pytest} is not intrinsically
good or bad --- the same command is decisive under one hypothesis and
irrelevant under another --- so per-action credit is a noisy target that
rewards surface form. A \textbf{terminal episode label} is too coarse in
the opposite direction: a failed trajectory can hold several
well-grounded phases before one wrong commitment, and a successful one
can hold long detours before a late correction, yet a single bit covers
both. A \textbf{fixed window} shortens the context without fixing the
semantics, since a \(K\)-turn boundary can split one coherent
investigation or merge the abandonment of one cause with the adoption of
the next. In every case the boundary comes from logging mechanics rather
than from the agent's changing understanding of the task.

\subsection{Hypothesis lifecycles as temporal
abstractions}\label{hypothesis-lifecycles-as-temporal-abstractions}

Coding-agent search is often organized around a causal hypothesis. The
agent adopts an explanation, gathers evidence relevant to it, modifies
code or runs a discriminating check, and then either retains or changes
the explanation. The observable lifecycle is

\[
\text{adopt }h_k \rightarrow \text{gather evidence} \rightarrow \text{edit/check}
\rightarrow \{\text{adopt }h_{k+1},\ \text{trajectory end}\}.
\]

We call the interval governed by one hypothesis a \textbf{semantic
phase}. A phase may span many tool calls or only a few; its length is
determined by the persistence of a causal conjecture, not by a message
count. In our collection protocol the acting agent declares each
adoption event, so phase start and end are observable through successive
declarations or trajectory termination. This is \textbf{semantic
self-segmentation during collection}: the boundary comes from the
policy's own epistemic commitment under a domain-general declaration
protocol, rather than from task-specific boundary labels or a separate
post-hoc segmenter.

Each phase also defines a local epistemic task. A \texttt{grep}, file
read, edit, or test is not an isolated action but an attempt to gather
evidence for, discriminate, or act on the phase's governing hypothesis.

\subsection{What a declaration costs, and what everything else
costs}\label{what-a-declaration-costs-and-what-everything-else-costs}

A declaration is cheap. Across the 876 trajectories of the ten
collection runs that carry one, the agent emitted 4,464 declarations at
a mean of 52 tokens: 233,601 tokens against 2,391,128 the agent
generated, so the protocol adds \textbf{9.8\% to what the agent writes}.
Against the whole transcript, which tool output dominates and the
protocol cannot influence, the same declarations are 0.57\%. We quote
both because the first is what the protocol costs and the second what it
occupies. Token counts are a plain-content character estimate, not a
tokenizer run, released as an artifact.

That matters because of what the alternatives require instead. Every
nearby method buys its local unit with something the collection process
does not have on hand: a gold developer patch, an environment-specific
milestone vocabulary, a replayable environment, white-box teacher
logits, or a second model that segments the trace after the fact. A
declaration protocol costs one instruction in the acting agent's prompt,
and yields the boundary inside the same rollout that produces the
trajectory. Table F1 sets this out against the closest published
alternatives. The price is that the protocol must be in place before
collection starts, so the method does not apply to logs already recorded
without it.

That is a claim about dependencies, not quality, and it is worth nothing
unless a unit obtained this cheaply is a real one.

\subsection{Contributions}\label{contributions}

A declared boundary is trivially retrievable from the log that declared
it, and that fact is worth nothing: it shows only that we recorded what
we recorded. What decides whether the unit is real is what \textbf{an
observer who never sees the declaration} can recover. Four contributions
follow from asking that.

\begin{enumerate}
\def\labelenumi{\arabic{enumi}.}
\tightlist
\item
  \textbf{A boundary that costs no external dependency.} The acting
  agent supplies the segmentation itself, under a declarative contract
  that names the unit to maintain, the event exposing its boundaries,
  and the criterion for auditing a completed unit. Nothing else is
  consulted: no gold developer patch, no environment-specific milestone
  vocabulary, no replayable environment, no white-box teacher logits,
  and no second model segmenting the trace afterwards. The boundary
  arrives inside the same rollout that produces the trajectory, which is
  what the alternatives cannot do
  \citep{ma2026patches, wang2026beacon, li2026gear, carl2026, xiong2025stepwiser}.
  We claim the specific combination --- a \emph{falsifiable causal
  hypothesis, declared by the acting agent during a tool-using rollout,
  used as a training-data boundary} --- and, since all three corpora are
  code-agent tasks, claim task-agnosticism only there.
\item
  \textbf{A unit other methods can attach to, because it has a name.} A
  window or an episode can only be pointed at by position; a declared
  hypothesis can be pointed at by what it asserts, which makes a phase
  an addressable handle rather than merely a shorter span. Several
  existing lines of work need such a handle and currently have to
  manufacture one. \emph{Finer-grained credit}: phase-level advantage
  replaces one episode return with one verdict per causal attempt
  \citep{zhang2026credit, luo2025agentlightning}, and because a boundary
  is a deployment-faithful state, it is also a legitimate resampling
  anchor --- \(G\) continuations from the same declared decision point
  form a group whose returns are comparable because they answer the same
  question \citep{shao2024deepseekmath}; Appendix E defines such an
  assigner over our verdicts. \emph{Process supervision without step
  annotation}: process reward models need a label per step, which is
  where their annotation cost sits \citep{lightman2023verify}, whereas
  here the step is the phase and the retrospective audit already
  attaches a verdict and a diagnosis to 2,721 of them.
  \emph{Segment-level preference learning}: methods that contrast at a
  segment rather than a turn or a session must select that segment
  retrospectively \citep{kong2025sdpo, chen2026agenticdpo}; here it is
  given. \emph{Guidance and elimination by name}: the eliminated set
  accumulates in the agent's own vocabulary, so a critic can negate one
  conjecture rather than a whole trajectory, and the record stays
  legible afterwards as which direction was ruled out and why.
  \emph{Context management}: a completed causal attempt is a
  summarisation boundary in a way a \(K\)-turn window is not, since what
  may be dropped is decided by whether the question it asked was
  settled. We instantiate the fourth of these (Section 3.1) and consume
  the third (Section 6); the rest are what the interface admits, not
  results we claim.
\item
  \textbf{The cuts follow the agent's reasoning, and are not an artefact
  of the logging.} Two things have to hold together for the unit to be
  worth anything, and we test both with every declaration scrubbed. It
  is not arbitrary: an observer that never sees a declaration still
  recovers a substantial part of the structure --- a model attributes
  action blocks to their governing hypothesis at over twice chance,
  beats equal-length windows over the same trajectories under a paired
  sign test, survives a model-free lexical control, and collapses under
  a label permutation; a code-blind human-protocol annotator places
  boundaries far above random. That much recoverable signal is what
  shows the boundary tracks the behaviour rather than the format. And it
  is not free: no observer reproduces the boundaries reliably, and a
  mechanical execution-event rule finds nothing at either end of a
  strict-to-permissive sweep, so the segmentation is not something a
  parser over the same logs would have supplied. Legible but not
  derivable is precisely the regime in which asking the agent is worth
  doing.
\item
  \textbf{A downstream test of the unit as a preference boundary.} We
  construct 2,551 phase-boundary preference pairs from masked
  deployment-faithful states and fit a standard optimizer to them,
  evaluating on two held-out sets built to be adversarial and on
  matched-construction items with two controls. Section 6 reports what
  we find.
\end{enumerate}

\section{Related Work}\label{related-work}

BEACON is the closest neighbour, partitioning long-horizon trajectories
at task-meaningful environment milestones with dual-scale advantage
estimation \citep{wang2026beacon}; like ours it rejects uniform windows
for variable-length phases. Appendix I places both in the wider
literature on temporal abstraction, agentic credit granularity, and
segment-level preference units.

Closer to our collection protocol, HHD escalates failed rollouts to a
teacher that sees the reference solution and distils its hint into a
scaffolded retry \citep{wang2026hhd}, P2T withholds the gold patch while
steering along milestones derived from it \citep{ma2026patches}, and
HDSO accumulates refuted hypotheses as negative evidence --- proposed by
a curator observing traces after the fact \citep{hdso2026}. CARL
segments tool-use rollouts at structural rather than semantic delimiters
\citep{carl2026}.

Two differences run through all of them. The \textbf{boundary author}
here is the acting policy, exposing its currently adopted causal
explanation as it works, so no milestone vocabulary has to be written
per environment, and a reviewer has something the agent itself asserted
to refuse. And these methods segment trajectories they are given,
whereas the collection protocol of Section 3.1 shapes which trajectories
exist.

Concurrently and on a different task morphology, Tycho has a coding
agent author its own revisable hypothesis about an interactive
environment --- as an executable program rather than a causal sentence
--- and decide whether to construct, repair, query, plan through, or
bypass it \citep{lehmann2026tycho}. The organising loop is the same as
ours: the agent commits to a conjecture, predicts, meets the
environment's verdict, and revises. We read the convergence as evidence
for the abstraction rather than against either instantiation, and take
one of their results as motivation: repairing a falsified model
automatically improves how well it reproduces observed transitions yet
plays worse than letting the actor request a rebuild, which is the
failure mode our escalation ladder avoids by retiring a refuted
hypothesis instead of patching it. Appendix I gives the comparison and
the caveats.

Appendix F tabulates the dependency claim of Section 1.3 against these
methods; it is about what each boundary costs to create, not empirical
dominance. One qualification: retrospective audit and preference
synthesis still use model judgments, so the pipeline is not ``LLM-free''
--- but those components evaluate an already exposed causal unit rather
than inventing its boundary after the fact.

\section{Method}\label{method}

\subsection{How the trajectories are
made}\label{how-the-trajectories-are-made}

Prior segmentation work takes trajectories as given. We generate them,
under an escalation designed to produce the transitions the unit is
meant to capture. An agent works the task alone; if it submits and the
hidden tests fail it receives the failure and tries again; only if it
still fails does a reviewer model intervene. That reviewer can see the
reference fix but must return exactly two things --- a negation of what
is wrong in the current approach, and the smallest positive direction to
investigate next --- and is forbidden from naming the fix, a file or
line, any identifier taken from the reference, or any code. Refuted
hypotheses accumulate across rounds and are carried forward as
eliminated, so the agent is pushed into a new region rather than allowed
to circle. The three regimes are recorded as \texttt{self\_solved},
\texttt{self\_corrected}, and \texttt{oracle\_redirected}.

\textbf{What the declaration adds is not the guidance but its address.}
What an undeclared trajectory does not supply is a proposition to
negate: the reviewer would have to infer what the agent is currently
attempting before it could refuse anything specific. With a declaration
it negates a named causal claim, adds it to an accumulating eliminated
set, and steers the agent out of that region by name. Neither the
escalation nor the eliminated set is novel in itself (Section 2); whose
hypothesis is being refused is. We should be plain about what this does
not establish: an eliminated set could instead be accumulated over
attempted patches or touched files, and we did not run the reviewer
without declarations, so the claim is that the declaration makes the
attachment point explicit and free, not that no substitute exists.

The same addressability survives into training. A refuted hypothesis is
carried forward verbatim with the diagnosis of what was wrong, so a
later target is conditioned on \emph{which} direction was eliminated and
\emph{why}, rather than on an undifferentiated record of failure.
Recorded solutions rarely contain explicit wrong-cause-then-correction
moments; here every redirect forces one against a conjecture verified
wrong, and the phase boundary is what keeps the refuted span separable
from the corrected one. Without it a redirected trajectory could only be
kept whole, importing the reviewer's influence into every action, or
discarded. The guidance text itself is masked out of every
deployment-faithful context: it appears in 0 of the 6,911 candidate pair
prompts the scan covered. Preference assembly reads a narrower pool of
5,858 candidates and retains 2,551 after recorded raw-leak (107),
hidden-token leakage (89), duplicate (764), holdout (19),
per-instance-cap (2,194), and global channel-share (134) filters; the
SFT export reads 4,971 rows and retains 4,679 after removing 271 old
raw-leak rows and 21 above its character-estimated window.

\subsection{Phase extraction}\label{phase-extraction}

Table H1 in Appendix H shows every phase of one trajectory. A
declaration opens a phase, and actions and observations stay attached to
it until the next adoption or the end of the trajectory; extraction
stores the edits, tests, locations, and read evidence bound to that
span, and a trace without a usable declaration is marked unbound rather
than counted as a clean phase. The declaration fixes the semantic object
and the auditor then evaluates the work performed under it, so contract,
boundary event, extractor, and validity criterion all refer to the same
thing. Reproducing the literal declaration string is no part of the
claim: the declaration is instrumentation for construction, and Section
5 measures the unit with every declaration scrubbed. A different
contract would define a different unit, obtainable by recollecting
rather than by re-scoring this data (Appendix J).

\subsection{Phase record and audit}\label{phase-record-and-audit}

A phase record contains the trajectory prefix available before the
decision, the governing hypothesis, the actions and evidence bound to
it, the boundary and retrospective status, and the audit verdict. This
record is a \textbf{local experience unit}: it retains enough causal
context to explain why an action sequence was attempted, while excluding
unrelated earlier and later directions.

The audit pass synthesizes \texttt{REASONING}, \texttt{CHECK}, and
\texttt{GAP} from the completed trace, flagging recurring failure modes:
an assumption already contradicted by available evidence, a direction
that does not discriminate the proposed cause, weak validation of a
plausible repair, repetition of a refuted direction, and a premature
conclusion. These are retrospective construction annotations, not
deployment observations, and not ground truth in every case.

\subsection{What one instrumented collection
yields}\label{what-one-instrumented-collection-yields}

Table 1 sets these out, and Figure 6 in Appendix G traces one collection
through them. The trajectories are the agent's own: it works under the
declaration contract and is then left to search, with no human marking a
boundary and no gold patch consulted to place one. That single rollout
produces not one training set but four, because a completed phase can be
read from four angles.

\begin{table}[!ht]
\textbf{Table 1: Four supervised targets from the same collection.}

{\def\LTcaptype{none} 
\begin{center}
\begin{tabular}{@{}
>{\raggedright\arraybackslash}p{(\linewidth - 6\tabcolsep) * \real{0.1802}}
  >{\raggedright\arraybackslash}p{(\linewidth - 6\tabcolsep) * \real{0.3874}}
  >{\raggedright\arraybackslash}p{(\linewidth - 6\tabcolsep) * \real{0.3784}}
  >{\raggedleft\arraybackslash}p{(\linewidth - 6\tabcolsep) * \real{0.0541}}@{}}
\toprule\noalign{}
\begin{minipage}[b]{\linewidth}\raggedright
Target
\end{minipage} & \begin{minipage}[b]{\linewidth}\raggedright
Sees
\end{minipage} & \begin{minipage}[b]{\linewidth}\raggedright
Learns to
\end{minipage} & \begin{minipage}[b]{\linewidth}\raggedleft
Rows
\end{minipage} \\
\midrule
audit & the completed phase and its bound evidence & judge whether a
direction was sound, and name what was missing & 2,721 \\
propose & only what was visible at the boundary & derive a checkable
next hypothesis from the facts, or state what evidence is missing (25
probe rows) & 1,405 \\
fix & an outcome-confirmed cause and the source & turn a localized
diagnosis into a patch & 553 \\
preference pairs & only what was visible at the boundary & prefer an
advancing opening over a stalling one & 2,551 \\
\bottomrule
\end{tabular}
\end{center}
}
\end{table}

\textbf{The information each target sees is deliberately asymmetric.}
Audit may use the completed phase, and fix is given a cause the outcome
confirmed, because both are retrospective construction steps; propose
and the preference pairs get only the masked boundary state a deployed
policy would have. The intended chain is \emph{judge a direction, infer
the next one from facts, repair once confirmed}, and only the last two
are ever scored against deployment-faithful context. Audit is where
failed search stops being waste: 1,316 of its 2,720 phases come from
spans the auditor rates weak or wrong --- regions an episode label
cannot separate from the phases that worked beside them --- and each
still carries a usable lesson about why the direction was unsound, which
no terminal bit preserves.

\subsection{The unit, stated formally}\label{the-unit-stated-formally}

Let a coding-agent trajectory be \[
\tau = (o_1, a_1, e_1, o_2, a_2, e_2, \ldots, o_T, a_T, e_T),
\] where \(o_t\) is the visible task and repository context, \(a_t\) is
a model action or tool invocation, and \(e_t\) is the resulting
observation, execution output, patch state, or test result. A terminal
outcome may be available after the full trajectory, but it does not
identify which intermediate search directions were useful.

A hypothesis \(h_k\) is the agent's current checkable conjecture about
the cause of the problem, the search region containing it, or the
mechanism a repair must change. It is valid when repository inspection
or execution evidence can support or refute it; it is not a claim that a
patch is already correct, and not merely a plan to read a file.

A semantic phase \(\phi_k\) opens when the agent adopts \(h_k\) and
contains the actions and evidence directed by it: \[
\phi_k = (x_k, h_k, A_k, E_k, b_k),
\] with \(x_k\) the trajectory prefix before the phase, \(A_k\) its
action sequence, \(E_k\) the bound evidence, and \(b_k\) the observable
closing boundary --- adoption of the next hypothesis, or trajectory
termination. Support, refutation, supersession and abandonment are
retrospective readings attached when the evidence permits; the corpus
does not store an exhaustive boundary-type field, which is why
termination coverage is zero rather than merely low.

The definition is semantic but observable: adoption events surface as
declarations, and the auditor binds the surrounding trace to the
declared causal object. Validity does not require monotone motion toward
the answer. Repository orientation, dependency inspection, failed
commands and reasonable detours stay inside the same phase while the
governing conjecture is unchanged, and refutation is progress as much as
support is --- it eliminates a live region. A refuted phase is a valid
experience unit even when the repair appears only in the next one.

\subsection{What is optimized, and what merely scopes
it}\label{what-is-optimized-and-what-merely-scopes-it}

The phase and the optimized span play different roles. The phase is the
\textbf{experience and credit unit}: its actions, evidence and outcome
reveal retrospectively whether its governing hypothesis advanced the
task. The optimized completion is the \textbf{decision at the boundary}
--- a compact statement of the hypothesis that would open the next
phase. Phase-DPO therefore never asks the model to reproduce a recorded
phase. Given the masked state \(x\) at a decision point, we seek \[
\pi_\theta(y^+ \mid x) > \pi_\theta(y^- \mid x)
\] for a preferred opening \(y^+\) over a rejected one \(y^-\). The
comparison is local: it asks which next conjecture has positive progress
potential from the same state, and that conjecture in turn fixes the
downstream search region. Plausible wording is not sufficient --- the
hypothesis must open a checkable phase expected to remove uncertainty,
discriminate live alternatives, or enter a repair-relevant region. The
lifecycle supplies the delayed evidence that labels the decision; the
actor sees only what was available at the boundary.

\subsection{Why the opening, and not some other
span}\label{why-the-opening-and-not-some-other-span}

Four properties pick out the opening. It carries decision leverage: one
conjecture organizes many later reads, probes, edits and tests, so
improving it can redirect a coherent stretch of search without pricing
each tool call on its own. It is credit-aligned, because evidence
accumulated across the completed phase bears on the hypothesis that
governed it, which is delayed supervision for the earlier decision
rather than for an unrelated one. It is comparable, since two candidate
openings can be scored against the same repository state and the same
prefix, where two complete phases would differ in tool results,
intermediate state and length at once. And it is what a deployed policy
actually controls: at inference the model can choose a conjecture but
not the command outputs that follow, so the boundary decision is the
controllable part.

The alternatives fail in identifiable ways. A next-action target is too
local, because the value of a search or a test depends on the active
causal theory. A full-phase target mixes the policy's choices with
externally determined tool feedback and reintroduces a length confound.
A final-patch or whole-trajectory target restores exactly the coarse
credit the segmentation was introduced to avoid. This also fixes what
the preference means: the relation is \emph{can advance the search}
rather than \emph{reads as more complete}, so a rejected opening is one
that stalls, repeats a refuted direction, or commits to the wrong
region.

One consequence is worth stating plainly, because it bounds what this
arm is. The objective is a contextual bandit over a single decision ---
no state transition, no advantage term. The phase enters it only as the
thing that fixes where context is cut and which text is the target, not
as the interval over which advantage is normalized. That second use is
the one the boundary was designed for, and it is not what is trained
here; this is neither a new objective nor an implementation of
semantic-phase group-relative optimization.

\section{Corpus}\label{corpus}

The corpus is real tool-using software-agent trajectories from SWE-style
repository tasks, LiveCodeBench, and Terminal-Bench: across ten source
runs, 904 trajectory artifacts over 790 task instances and 122
repositories or suites.

\begin{table}[!ht]
\textbf{Table 2: Trajectory and phase corpus statistics.} Phase length
is given as messages / tool calls / plain-content tokens, the last a
character-based estimate released as an artifact rather than a tokenizer
run.

{\def\LTcaptype{none} 
\begin{center}
\begin{tabular}{@{}
>{\raggedright\arraybackslash}p{(\linewidth - 2\tabcolsep) * \real{0.4286}}
  >{\raggedleft\arraybackslash}p{(\linewidth - 2\tabcolsep) * \real{0.5714}}@{}}
\toprule\noalign{}
\begin{minipage}[b]{\linewidth}\raggedright
Quantity
\end{minipage} & \begin{minipage}[b]{\linewidth}\raggedleft
Value
\end{minipage} \\
\midrule
source task instances / coarse repositories & 790 / 122 \\
raw trajectories & 904 \\
trajectories with ledgers / with retrospective audits & 852 / 803 \\
retained audit keys / spans reconstructed & 2,721 / 2,720 \\
phases per contributing trajectory, mean / median & 3.59 / 2 \\
phase length, median & 16 / 9.5 / 8,892 \\
phase length, p90 & 61 / 37 / 44,008 \\
preference pairs (outcome-verified / LLM-judged) & 2,551 (726 /
1,825) \\
\bottomrule
\end{tabular}
\end{center}
}
\end{table}

\textbf{The declaration covers the behaviour it is supposed to bound.}
Table 2 gives the counts. Of 2,721 retained audit keys, 2,720
reconstruct a valid span; the single failure is a ledger pointing at
\texttt{msg\#45} in a 41-message trajectory. Every reconstructed span
contains a declared hypothesis and only 2 contain more than one distinct
normalized hypothesis, so the boundary is not silently merging
directions. Only 1 of 45,587 tool calls in contributing trajectories
precedes the first declaration --- calls after a phase opens are bound
by construction, so this measures how early the agent declares rather
than coverage per se. Spans are genuinely variable: a median of 16
messages and 9.5 tool calls against a 90th percentile of 61 and 37.
These are deterministic checks on coverage and binding, not on semantic
correctness.

\textbf{A blinded annotator finds the spans usable.} One independent
model annotator worked through a fixed-seed worksheet of 50 retained
phases, five per source run, before the automatic-audit key was opened.
It rates 48 usable as local experience units (96.0\%; Wilson 95\% CI
86.5--98.9\%), with 49/50 valid starts, 49/50 single-hypothesis
coherence, 50/50 binding actions and evidence to the governing
hypothesis, 49/50 valid ends and 50/50 free of future-information
leakage. Both rejects are real boundary errors rather than borderline
calls: one declaration restates rather than terminates the active
lifecycle, and one phase opens with a broad search program over several
candidate modules instead of a causal conjecture. The annotator also
assigns 9 refute, 20 supersede, 2 support, 18 terminal and 1 uncertain
endpoint labels. This is a model-assisted audit rather than human
evaluation, and it samples only retained phases.

\textbf{Leakage wall.} Export refuses to build a training package unless
a frozen instance holdout is present, and the held-out sets screen more
widely still: an instance is eligible only if it contributes no pair to
either channel. The final export metadata lists 501 training instances
with zero overlap against the frozen split; the held-out sets screen
against all 538 instances appearing in either the natural or the
resample channel and admit 175 of the 789 carrying both a ledger and an
audit. Phases from one trajectory are never split between train and
validation.

\section{Do the declared segments hold up under
inspection?}\label{do-the-declared-segments-hold-up-under-inspection}

Reconstruction shows a declared boundary can be retrieved from the
ledger; it does not show that anything happens there. We therefore ask
two questions of observers who never see the declaration, and keep them
apart because they need different tests and support claims of different
strength. First: is a phase internally \textbf{coherent} --- are the
actions inside it attributable to the hypothesis that governed them?
Second: could a \textbf{cheaper rule} have produced the same boundaries?

\textbf{Setup.} Every test removes the declaring message and scrubs
every hypothesis string from the remaining text, so what is scored is
only what the agent subsequently did. This scrubbing is not cosmetic:
leaving the declaring message in place lets the hypothesis be found by
string match --- an unscrubbed version of the attribution test scored
0.576 that way --- so every declaring message is deleted and every
hypothesis string is redacted from what remains. The two
boundary-placing observers are not told how many boundaries exist; the
attribution test in Section 5.1 is given the count by construction,
since it is handed \(K\) blocks.

\subsection{Actions are attributable to their governing
hypothesis}\label{actions-are-attributable-to-their-governing-hypothesis}

A trajectory declaring \(K\) hypotheses is cut into \(K\) action blocks;
each block is shown with all \(K\) hypotheses from that trajectory and
must be matched to its own. Chance is \(1/K\), and because the
distractors are siblings, repository, task, and vocabulary are
controlled by construction. Table 3 collects the attribution results.
Over 221 phases from 49 trajectories with \(3 \le K \le 12\), an
independent model reaches 2.18 times chance (95\% trajectory-clustered
lift {[}+0.180, +0.336{]}), and accuracy stays roughly flat as chance
falls with \(K\) --- 0.485 at \(K{=}3\) against 0.333, 0.542 at
\(K{=}12\) against 0.083.

Two controls bound the reading. A model-free TF-IDF cosine with
within-trajectory IDF reaches 0.353 on the identical items, so roughly
half the effect survives without any model judgment. A label permutation
on a second, larger sample collapses accuracy to chance. Figure 3 shows
what happens when every boundary is slid instead: attribution decays
away from the declaration in both directions, but the maximum sits two
messages \emph{earlier}, at 0.271 against 0.219. We report no interval
at any offset and did not test that difference, so we take from the
curve only that position matters, not that the declared position is
optimal: a fixed two-message backshift is a placement rule we have not
ruled out. Appendix B gives both control samples in full.

\begin{table}[!ht]
\textbf{Table 3: Attribution accuracy, with the declaration deleted and
the hypothesis text scrubbed.} The middle two rows come from one paired
run in which a single matcher scored both segmentations of the same
trajectories, so any bias it has cancels between them. Blocks under 200
rendered characters are skipped in both arms; this drops more declared
than window blocks --- short declared phases are the hardest to
attribute --- so the filter runs in the declared arm's favour and the
row counts differ.

{\def\LTcaptype{none} 
\begin{center}
\begin{tabular}{@{}
>{\raggedright\arraybackslash}p{(\linewidth - 8\tabcolsep) * \real{0.4111}}
  >{\raggedleft\arraybackslash}p{(\linewidth - 8\tabcolsep) * \real{0.1778}}
  >{\raggedleft\arraybackslash}p{(\linewidth - 8\tabcolsep) * \real{0.1111}}
  >{\raggedleft\arraybackslash}p{(\linewidth - 8\tabcolsep) * \real{0.0889}}
  >{\raggedleft\arraybackslash}p{(\linewidth - 8\tabcolsep) * \real{0.2111}}@{}}
\toprule\noalign{}
\begin{minipage}[b]{\linewidth}\raggedright
Blocks scored
\end{minipage} & \begin{minipage}[b]{\linewidth}\raggedleft
Blocks / traj.
\end{minipage} & \begin{minipage}[b]{\linewidth}\raggedleft
Accuracy
\end{minipage} & \begin{minipage}[b]{\linewidth}\raggedleft
Chance
\end{minipage} & \begin{minipage}[b]{\linewidth}\raggedleft
\(\times\) chance
\end{minipage} \\
\midrule
declared & 221 / 49 & 0.484 & 0.222 & 2.18 \\
declared, paired run & 1884 / 374 & 0.495 & 0.208 & \textbf{2.38} \\
equal-length, paired run & 1917 / 374 & 0.431 & 0.208 & 2.07 \\
lexical control, no model & 221 / 49 & 0.353 & 0.222 & 1.59 \\
\bottomrule
\end{tabular}
\end{center}
}
\end{table}

\textbf{Would an arbitrary cut do as well?} The test above supplies the
cut points, so it measures coherence rather than placement, and
coherence is only interesting if the declared boundary produces more of
it than a mechanical one. We therefore ran a paired experiment over 374
trajectories: one matcher scored both the declared blocks and, on the
same trajectories with the same hypothesis lists, \(K\) equal-length
blocks spanning the same messages. A window straddling two hypotheses is
scored against whichever governs more of it, so the window arm is given
the benefit of the doubt. Declared blocks reach 0.495 against chance
0.208 (2.38x, 95\% trajectory-clustered {[}2.25, 2.51{]}); equal-length
windows reach 0.431 against 0.208 (2.07x, {[}1.96, 2.20{]}). Their
marginal intervals are disjoint: pairing by trajectory, the declared arm
is better on 162 and worse on 100 of the 262 that separate them (sign
test \(p = 0.0002\)), with 112 tied; Figure 4 in Appendix B shows the
per-trajectory differences.

The lexical control makes the reason explicit. TF-IDF scores 0.316 on
declared blocks and 0.305 on equal-length ones --- it cannot tell the
two segmentations apart, while the model can. On declared blocks the
model exceeds the lexical baseline by +0.178 (95\% trajectory-clustered
{[}0.152, 0.206{]}). What the declared boundary adds is therefore not
vocabulary but a span whose actions cohere under one explanation.

\subsection{A reader who cannot read the
code}\label{a-reader-who-cannot-read-the-code}

The first author annotated 13 held-out trajectories seeing only the
commands issued and their outputs, with a machine-generated per-command
description as a reading aid, screened for any reference to cause,
intent, or transition. The annotator lacks engineering-level familiarity
with these repositories and worked from behavioral pattern alone. This
is an interested annotator, so we state the barrier exactly: they knew
the segmentation criterion, while the interface withheld the number of
boundaries, their positions, and the hypothesis text. Neither the mark
budget nor the primary statistic was pre-registered, and the residual
bias runs toward over-recovery.

Against 40 declared boundaries the annotator matched 24 within one step,
where randomly placing the same number of marks matches 11.5
(permutation test, \(p < 0.0001\)). The cost is precision: 62 marks were
placed for those 24 hits, a 1.55-fold over-segmentation consistent with
a reader who cannot tell a change of file from a change of explanation.
This is a floor, not what an expert annotator would achieve.

\textbf{Metrics and multiplicity.} Under the standard segmentation
metrics the same annotation looks unimpressive: \(P_k\) is 0.452 against
0.488 for random marks at the same count, and WindowDiff is 0.484. We
report both for comparability, but they penalise over-segmentation
heavily enough to mask the matching signal a 1.55-fold over-segmenter
produces, which is why the permutation test on matches is the primary
statistic; that choice was made after the metrics were computed and was
not pre-registered. Four of the 13 trajectories are individually
significant at \(p < 0.05\), including one exact 5-of-5 match, but the
13 tests are uncorrected, so roughly one significant result is expected
by chance. Post hoc, those four appear to be the ones organised around a
visible test-fix-test loop while the rest require repository
understanding; we did not pre-specify that split and offer it as
description rather than finding.

\subsection{A mechanical rule does not reproduce the
boundaries}\label{a-mechanical-rule-does-not-reproduce-the-boundaries}

The natural mechanical competitor for a repair-oriented trajectory is
the test-outcome event: cut after every test-suite invocation. Its yield
depends entirely on how broadly ``test'' is read, so Figure 1 reports
both ends, and neither beats chance. They fail in opposite ways: the
strict rule is too sparse to find the boundaries, and the permissive one
matches them only by covering most of the trajectory, at 10.6\%
precision. The failure is therefore a property of the rule family rather
than of one threshold. Declared boundaries do not coincide with test
events, even though the trajectories a code-blind reader can follow are
precisely those organized around a test-fix-test loop: what makes those
trajectories legible is not what marks the boundary. Appendix D gives
both rules.

\begin{figure}
\centering
\includegraphics[width=0.74\linewidth,height=\textheight,keepaspectratio,alt={A code-blind reader finds boundaries; a mechanical rule with the same access does not. Boundaries matched out of 40, against what the same number of randomly placed marks would match on the same trajectories. Neither the sparse nor the permissive test-event rule beats its own chance baseline, at either end of the sweep. No observer is told how many boundaries to expect.}]{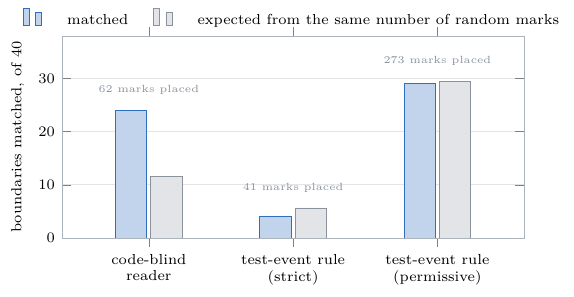}
\caption{A code-blind reader finds boundaries; a mechanical rule with
the same access does not. Boundaries matched out of 40, against what the
same number of randomly placed marks would match on the same
trajectories. Neither the sparse nor the permissive test-event rule
beats its own chance baseline, at either end of the sweep. No observer
is told how many boundaries to expect.}\label{fig:placement}
\end{figure}

\subsection{What this establishes}\label{what-this-establishes}

Two things follow. A phase is a semantic object an outside observer can
check rather than a bookmark we wrote into our own log: its actions stay
attributable with the declaration deleted, more so than for equal-length
windows over the same trajectories, and attribution decays away from the
declared position. And the boundary is not what a cheap rule would
produce --- a test-event rule matches no more than random placement at
either end of its sweep, while the observer that does find positions
buys them by proposing 62 marks for 24 hits.

What we do not claim: nothing here recovers boundaries from an
undeclared log --- of the three observers, one was handed the cut
points, one over-segments and one fails --- and nothing shows a policy
trained on these units is better. Section 9 states both.

\section{Downstream consumption}\label{downstream-consumption}

A phase is a credit unit before it is a preference unit: as a
\textbf{preference unit} it fixes where the context is cut and which
text is the target; as a \textbf{credit unit}, the interval over which
advantage is normalized. Neither is this paper's contribution. What this
section reports is what one standard optimizer does with the pairs, and
what the two held-out sets say about where the limit currently sits.

\textbf{As a preference unit.} Standard DPO consumes the 2,551 pairs
against a merged 27B SFT policy with a fresh LoRA adapter, objective
unmodified. Training separates in an orientation-sensitive way a
label-inverted arm does not reproduce (final reward margin 1.2133
against \(-0.0027\)) --- an optimization result only; Figure 5 in
Appendix C shows why nothing flips. On 91 adversarially built held-out
items over 87 never-trained instances, all four arms score exactly 0.824
and \textbf{no adapter changes a single decision} (Table 4). The adapter
is not inert: its median shift is 1.40 against the label-inverted arm's
0.37, so the orientation is specific rather than a generic perturbation.
But it is far too small and directionally random --- 1.40 against a
median candidate gap of 18.89, direction 0.505 --- so the null cannot
distinguish a preference that failed to transfer from one that was never
there. A 20-task rollout pilot resolves 13, 14 and 14 tasks for base,
SFT and Phase-DPO.

\textbf{The null is construction-bound.} Those items are adversarial ---
the rejected side is a detour written to compete with the chosen
hypothesis, not drawn from the channel that produced the training pairs.
Since 1,825 of the 2,551 pairs come from one resampling family --- three
pair types built by the same generator --- we built a \emph{matched}
set: 60 items over 15 never-trained instances from that same family,
every instance screened against the 538 contributing to either training
channel. There the same adapter moves four decisions, all wrong to
right, while both controls move none, and its per-item direction is
0.617 against the 0.505 it showed on adversarial items. We do not test
that difference, and the two sets differ in construction, instance pool,
size and difficulty at once; exact McNemar gives \(p = 0.125\) with the
other three tests between 0.09 and 0.13. This is \textbf{consistent
with} construction-bound fitting rather than proof of it, and the next
step is a more diverse pair corpus, not a different boundary. Appendix C
gives every test.

\begin{table}[!ht]
\textbf{Table 4: The same three adapters on two held-out sets.} Both
hold out whole instances that contribute no training pair. They differ
in construction, but also in instance pool, item count and baseline
difficulty, so construction is not isolated.

{\def\LTcaptype{none} 
\begin{center}
\begin{tabular}{@{}
>{\raggedright\arraybackslash}p{(\linewidth - 4\tabcolsep) * \real{0.6087}}
  >{\raggedleft\arraybackslash}p{(\linewidth - 4\tabcolsep) * \real{0.1884}}
  >{\raggedleft\arraybackslash}p{(\linewidth - 4\tabcolsep) * \real{0.2029}}@{}}
\toprule\noalign{}
\begin{minipage}[b]{\linewidth}\raggedright
\end{minipage} & \begin{minipage}[b]{\linewidth}\raggedleft
adversarial
\end{minipage} & \begin{minipage}[b]{\linewidth}\raggedleft
matched
\end{minipage} \\
\midrule
held-out items / instances & 91 / 87 & 60 / 15 \\
SFT reference accuracy & 0.824 & 0.700 \\
main-arm accuracy & 0.824 & \textbf{0.767} \\
decisions changed by the main arm & \textbf{0 / 91} & \textbf{4 / 60} \\
decisions changed by either control & 0 / 91 & 0 / 60 \\
base margin, median & 18.89 & 14.56 \\
main-arm shift, median / max & 1.40 / 7.87 & 4.05 / 40.62 \\
\bottomrule
\end{tabular}
\end{center}
}
\end{table}

\textbf{As a credit unit, in principle.} The consumer the boundary was
designed for normalizes advantage \emph{within} phases rather than
across trajectories, as BEACON does with its milestones
\citep{wang2026beacon}. Its offline component is implemented and
non-degenerate: of 877 resolved episodes, 522 declare more than one
phase and 355 of those (68\%) contain phases of differing verdict class,
receiving \textbf{opposite-signed} advantages exactly where the episode
label gives every phase the same sign. A declared boundary is also a
state a sampler can return to: it marks a decision the policy actually
made, so \(G\) alternative openings can be drawn from the same context
and scored against each other --- the shape a group-relative method
needs and an arbitrary token offset does not provide. We did not run
that loop: near the solve-rate floor a uniformly unsuccessful group has
identically zero advantage, a property of the deployment rather than of
the boundary (Appendix E). Online phase-level optimization is future
work.

\subsection{An exploratory twenty-task rollout
pilot}\label{an-exploratory-twenty-task-rollout-pilot}

The workshop question --- does a policy trained on these units act
better? --- has one piece of direct evidence, a pilot deliberately sized
for orientation rather than significance. Under an identical scaffold,
task set, context budget, and rollout budget, the unadapted base
resolves 13 of 20 held-out SWE-bench Verified tasks; SFT and Phase-DPO
each resolve 14. Eleven tasks are resolved by all three arms, the
post-trained arms share 12, two are SFT-only, two are DPO-only, and four
are solved by neither. The pilot establishes no resolve-rate
improvement, which is consistent with the construction-bound reading of
Table 4 rather than in tension with it.

\begin{table}[!ht]
\textbf{Table 5: Three-arm rollout outcomes and raw-trajectory
efficiency.} A message is one raw trajectory message object; a tool call
is one assistant tool-call object. The matched-success rows use only the
11 tasks resolved by all three arms, so early failure cannot
mechanically shorten a trajectory.

{\def\LTcaptype{none} 
\begin{center}
\begin{tabular}{@{}
>{\raggedright\arraybackslash}p{(\linewidth - 6\tabcolsep) * \real{0.6184}}
  >{\raggedleft\arraybackslash}p{(\linewidth - 6\tabcolsep) * \real{0.1053}}
  >{\raggedleft\arraybackslash}p{(\linewidth - 6\tabcolsep) * \real{0.1316}}
  >{\raggedleft\arraybackslash}p{(\linewidth - 6\tabcolsep) * \real{0.1447}}@{}}
\toprule\noalign{}
\begin{minipage}[b]{\linewidth}\raggedright
Metric
\end{minipage} & \begin{minipage}[b]{\linewidth}\raggedleft
base
\end{minipage} & \begin{minipage}[b]{\linewidth}\raggedleft
SFT
\end{minipage} & \begin{minipage}[b]{\linewidth}\raggedleft
Phase-DPO
\end{minipage} \\
\midrule
resolved tasks & 13/20 & 14/20 & 14/20 \\
mean messages, all 20 & 101.7 & 90.0 & \textbf{87.5} \\
message reduction vs.~base & -- & 11.5\% & \textbf{14.0\%} \\
mean tool calls, all 20 & 49.9 & 44.0 & \textbf{42.9} \\
tasks with fewer messages than base & -- & 11/20 & \textbf{13/20} \\
mean messages, all-three-resolved (n=11) & 102.2 & \textbf{87.2} &
87.4 \\
shorter than base, all-three-resolved & -- & 6/11 & \textbf{8/11} \\
\bottomrule
\end{tabular}
\end{center}
}
\end{table}

The efficiency movement is a tendency, not a stable gain: paired mean
message deltas against base are -11.7 for SFT (task bootstrap 95\% CI
{[}-29.5, 5.0{]}) and -14.2 for Phase-DPO ({[}-31.8, 4.5{]}), both
intervals containing zero. It persists on the 11 jointly resolved tasks,
so early failure is not the whole explanation, but that subset is small
and post hoc. Nor does the pilot separate Phase-DPO from its SFT
initialization: where both solve, DPO is shorter on seven of twelve,
paired mean -1.5 messages ({[}-15.0, 13.8{]}). We also refuse the
outcome-stratified reading, because conditioning on who solved a task
selects the arm that kept working until it succeeded --- measured that
way every stratum shows the winner longer, and the largest gap belongs
to the \emph{base} model. Table 5 is preliminary search-dynamics
evidence for the pipeline and nothing more.

\textbf{Four rollouts read individually.} The aggregate hides what
changed, so we describe four cases, chosen to cover improvement, an
unchanged outcome reached differently, and a regression; they are
descriptive and do not add aggregate evidence. On
\texttt{django\_\_django-11292} the SFT arm changes both the requested
command-line behaviour and the programmatic default, while the Phase-DPO
arm preserves the internal default and adds only the narrower behaviour;
DPO resolves and SFT does not. On \texttt{pydata\_\_xarray-2905} the SFT
arm tests array-likeness after extracting the underlying values, which
leaves the condition too broad, while Phase-DPO checks whether the
object itself exposes the required interface before extraction; again
DPO resolves and SFT does not. On \texttt{django\_\_django-16662} both
produce the same import-ordering fix, but SFT takes 104 messages and 51
shell actions against 60 and 29 --- part of that gap is less post-fix
verification, so we call it earlier convergence rather than better
reasoning. And on \texttt{psf\_\_requests-6028} the regression runs the
other way: SFT traces lost authentication to URL reconstruction and
resolves, while Phase-DPO commits to a proxy-header account in the wrong
locus and fails. A plausible commitment made early is exactly what a
preference over phase openings can get wrong.

\subsection{Setup, controls, and what each layer of evidence may
claim}\label{setup-controls-and-what-each-layer-of-evidence-may-claim}

Four evidence layers are kept apart and no metric is promoted across
them: representation quality, optimization diagnostics, held-out
preference generalization, and the exploratory rollout. They answer, in
order, whether declared hypotheses form reconstructable inspectable
units; whether the boundaries are independently recoverable and their
positions carry information; whether phase-derived pairs give an
orientation-sensitive signal that label inversion does not reproduce;
and whether that fitted preference survives held-out instances and a
change in how pairs are constructed.

All arms start from the same merged 27B SFT policy, which is both the
DPO reference and the behavioural baseline, so what is isolated is
preference optimization applied after the model has already learned the
audit, propose and fix views. The \textbf{label-inverted control} swaps
chosen and rejected on the same 2,551 contexts while holding
initialization, pair count and configuration fixed: if the main arm's
separation came from merely running the DPO shell, this arm would
reproduce it, and it does not. A \textbf{verified-only} arm trains on
the 726 execution-verified pairs alone; it separates too, ending at
margin 0.071 with training accuracy 0.719 and loss 0.6604, so the weaker
separation tracks the smaller pair count rather than the verification
channel.

\begin{table}[!ht]
\textbf{Table 6: Phase-DPO training configuration.} Identical across
arms except where shown; every value is read from the copied trainer
state rather than from the intended configuration.

{\def\LTcaptype{none} 
\begin{center}
\begin{tabular}{@{}
lrrr@{}}
\toprule\noalign{}
Setting & Main & Label-inverted & Verified-only \\
\midrule
pairs / optimizer steps & 2,551 / 319 & 2,551 / 319 & 726 / 91 \\
LoRA rank / alpha / dropout & 16 / 32 / 0.05 & same & same \\
beta / cutoff & 0.1 / 8,192 & same & same \\
batch x accumulation & 1 x 8 & same & same \\
learning rate / epochs & 5e-6 / 1 & same & same \\
optimizer / seed & fused AdamW / 42 & same & same \\
trainable / total parameters & 116,727,808 / 27.5B & same & same \\
\bottomrule
\end{tabular}
\end{center}
}
\end{table}

On the evaluation side, an item answerable without reading the evidence
measures nothing, so the surface channels we can close are closed and
the ones we cannot are reported. Each arm assigns an implicit reward
\(\log \pi_{\theta}(y \mid x) - \log \pi_{\mathrm{ref}}(y \mid x)\) to
two fixed candidates, so no policy generates and no teacher completion
is scored. Presentation order is randomized per item; both candidates
are written under the same specificity contract and neither may hedge;
pairs whose sides differ by more than 8\% in characters are dropped; the
label is re-derived by a judge that did not write the candidates. We
report a pick-the-longer-side baseline so residual length signal stays
visible, and a token-normalized accuracy because a summed
log-probability grows with length. Because both sides are model-written,
a residual generator signature remains possible; we record it as a
limitation rather than claim its absence.

\section{The boundary as a context-compression
rule}\label{the-boundary-as-a-context-compression-rule}

A memory that only accumulates defeats itself: the transcript grows
linearly in tool output, and an agent that keeps everything eventually
cannot afford to consult any of it. Keeping experience means compressing
it, and compression needs a unit --- something that decides, for each
stretch of work, what inside it was load-bearing. The ledger of Section
3.2 is already that compressed form; this section measures how good a
compression it is, locally and without a model, since both
representations sit on disk for every collected trajectory.

The rule follows from the boundary: within a phase keep the hypothesis
and its trigger, every state-changing command, every test with its
result, up to eight \texttt{file:line:} hits ordered so that
eventually-edited files come first, and source snapshots of edited
files; collapse the remaining read, grep, and list steps to a count.
Recent test-time scaling work instead compresses rollouts into
model-written summaries preserving salient hypotheses and failure modes
\citep{kim2026ttc}; a learned summarizer's omissions cannot be
enumerated, and ours can --- they are exactly the uncapped hits and the
collapsed reads.

\begin{table}[!ht]
\textbf{Table 7: Retention at a matched character budget, over the 876
collection trajectories.} Each baseline receives the byte budget the
phase record used on the same trajectory. The record is 6.5\% of the
transcript overall (median 6.8\%, 90th percentile 13.7\%): a median
transcript of roughly 31,600 tokens becomes a record of roughly 1,900,
and a 90th-percentile transcript of roughly 141,000 tokens becomes
roughly 9,700.

{\def\LTcaptype{none} 
\begin{center}
\begin{tabular}{@{}
>{\raggedright\arraybackslash}p{(\linewidth - 4\tabcolsep) * \real{0.5541}}
  >{\raggedleft\arraybackslash}p{(\linewidth - 4\tabcolsep) * \real{0.2027}}
  >{\raggedleft\arraybackslash}p{(\linewidth - 4\tabcolsep) * \real{0.2432}}@{}}
\toprule\noalign{}
\begin{minipage}[b]{\linewidth}\raggedright
At the record's budget
\end{minipage} & \begin{minipage}[b]{\linewidth}\raggedleft
edit commands
\end{minipage} & \begin{minipage}[b]{\linewidth}\raggedleft
test invocations
\end{minipage} \\
\midrule
phase record & 99.97\% & 100.0\% \\
same categories, no boundary (ablation) & 99.97\% & 100.0\% \\
recency tail & 20.2\% & 29.3\% \\
uniform stride & 10.3\% & 19.6\% \\
\bottomrule
\end{tabular}
\end{center}
}
\end{table}

\begin{figure}
\centering
\includegraphics[width=0.78\linewidth,height=\textheight,keepaspectratio,alt={What a positional rule costs in budget. A recency tail --- the simplest positional policy, and what plain truncation keeps when a trajectory overruns its window --- retains 20.2\% of edit commands at the record's budget, needs eight times that budget to pass 79\%, and still reaches only 97.4\% at sixteen times, by which point it is holding essentially the whole transcript. The record sits at one unit and 99.97\%.}]{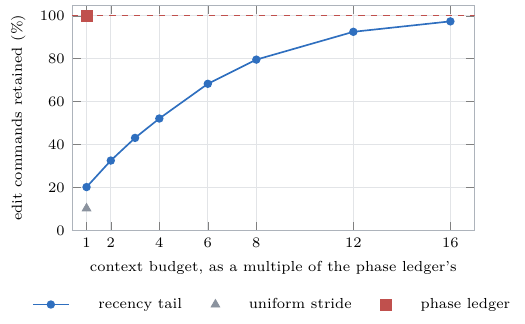}
\caption{What a positional rule costs in budget. A recency tail --- the
simplest positional policy, and what plain truncation keeps when a
trajectory overruns its window --- retains 20.2\% of edit commands at
the record's budget, needs eight times that budget to pass 79\%, and
still reaches only 97.4\% at sixteen times, by which point it is holding
essentially the whole transcript. The record sits at one unit and
99.97\%.}\label{fig:compression}
\end{figure}

The ablation row is reported against ourselves: the identical
keep-categories with \textbf{no phase structure at all} --- one global
bucket, one global eight-hit cap --- retain exactly as many edits and
tests in 79\% of the record's size. On retention alone the categories do
the work and the boundary is overhead. What it buys instead is
\emph{addressability}: each kept artifact arrives bound to the
conjecture that motivated it and the verdict that conjecture earned, the
structure every target of Section 3.4 consumes and a flat bucket does
not carry. The per-phase hit budget likewise pays only where there is
structure to exploit --- at four or more phases it covers an
eventually-edited file 69.9\% of the time against a global cap's 65.5\%,
at four times the kept hits.

What the compression drops is enumerable. It keeps 14,171 of 234,980 raw
search-hit lines (6.0\%), and the edited-file-first ordering makes the
kept set name an eventually-edited file in 67.7\% of the 744
trajectories where the question is defined, against 56.0\% for the same
number drawn uniformly (mean of 20 draws; paired difference 11.7 points,
95\% trajectory bootstrap {[}10.0, 13.5{]}) --- twelve points recovered
over an unordered cap, and still every pointer discarded in a third of
trajectories. Three of 9,444 edit commands are lost, all in one
trajectory that wrote brute-force verification scripts to a scratch
directory before declaring its first hypothesis; none touches repository
state. And this is retention of artifacts, not downstream performance:
no policy has been run with the record as its context, which is the
obvious next experiment.

\section{Discussion}\label{discussion}

\textbf{Why a phase is a plausible unit.} A phase carries more local
causal structure than a single action: searches, reads, edits and tests
execute under one hypothesis and can be read through the evidence
objective they serve. It mixes fewer conflicting directions than a whole
trajectory, and its length tracks the agent's epistemic state rather
than a fixed window. The binding checks support this reading, with the
qualification Section 4 already states: essentially every tool call in a
contributing trajectory falls after a phase has opened, but calls after
an opening are bound by construction, so the number measures how early
the agent declares rather than coverage as such.

The search-space contraction story behind the design is an intuition,
not a measured result. If a task's solution lies inside the agent's
reachable space, a useful phase should buy evidence that removes a live
alternative without discarding the correct region, and repeating such
updates over a bounded space may concentrate search on an answer. We
observe no scalar contraction function and prove nothing about local
improvements composing globally.

\textbf{What the training evidence establishes, and what it does not.}
Phase-level preferences produce an orientation-sensitive optimization
signal that a same-size label-inverted arm does not reproduce. That is
stronger than a falling loss curve and weaker than a capability claim:
on an independently constructed held-out set the fitted orientation is
simply unavailable, no decision changes, and the shift direction is at
chance. The matched-construction control locates what the separation was
-- on items built through the training pipeline's own channel the same
adapter changes four of sixty decisions where both controls change none
-- so the fitted preference is real but bound to how the pairs were
written. None of this establishes end-to-end coding capability, better
next-phase selection, or a higher resolve rate. A learned training
margin can coexist with no behavioural advantage, and a locally better
phase choice can still fail to produce a globally correct patch. Nothing
here observes whether a trained policy preserves the region containing
the correct cause: the held-out evaluation scores a choice between two
fixed candidates rather than a trajectory, and the rollout pilot scores
terminal outcomes rather than the search that produced them.

\textbf{Representation versus capability.} The contributions this paper
defends are the collection-time segmentation protocol and the data
interface built on it -- the acting agent exposes a variable-length
causal unit while generating the trajectory, and construction can then
retain useful work from failed episodes instead of stamping the terminal
label onto every action. Phase-DPO is the consumability result: an
existing optimizer trains on the unit and produces an
orientation-sensitive signal. It is not offered as an algorithmic
novelty, a mature DPO study, or a multi-seed capability result. The
rollout pilot adds a descriptive change in search dynamics and nothing
about \emph{adaptive} search-budget allocation: testing that would
require labelling how underdetermined each state is before any arm acts,
and sorting tasks by who eventually solved them makes the successful arm
longer regardless of policy.

\textbf{The consumer the unit was designed for.} A phase is a credit
unit before it is a preference unit. As a preference unit the boundary
fixes where context is cut and which text is the target; as a credit
unit it fixes the interval over which advantage is normalized and the
span that advantage reaches. The second use is what motivates
variable-length semantic phases in the first place. Its offline
component is implemented and non-degenerate: each phase takes a verdict
from the agent's own executed checks, a scalar reward whose
\emph{ordering} separates a surviving fix from one a later phase
overwrites and a clean refutation from a span reaching no verdict, and a
group-normalized advantage over the trajectory's phases. The magnitudes
are unvalidated design constants and no result depends on them; the
ordering is the design claim. A typical case is an episode recorded as
solved whose first hypothesis is cleanly refuted and whose second
survives: the two receive advantages of \(-1.0\) and \(+1.0\), where the
episode label gives both the same sign. Because mean-centring makes the
split depend only on the classes being distinct, the fraction of
episodes this affects is a property of the segmentation, not of the
magnitudes.

We do not close the loop, and the obstruction is the deployment rather
than the boundary. Three constraints compose. Group-relative
optimization must hold the trainable policy, a frozen reference, and
rollout KV cache on the device at once, where offline preference
optimization on the same hardware holds far less -- so the online policy
we could fit is materially smaller than the 27B this paper trains
offline. Scoring one repository rollout means executing a real test
suite in a container, tens of seconds to minutes per sample in our own
pipeline, against the thousands of samples an on-policy run consumes.
And decisively, the group-relative advantage is identically zero when a
sampled group is uniformly unsuccessful, so a policy near the solve-rate
floor on unseen repositories supplies no gradient at all -- an absent
update rather than a weak one. We did not benchmark the exact memory
ceiling or measure where that floor sits for a smaller policy, and quote
no figure for either; the argument is structural, since both constraints
push from the same side and no choice of boundary relaxes either. We
therefore trained no online arm rather than reporting an underpowered
one.

\section{Limitations}\label{limitations}

\begin{enumerate}
\def\labelenumi{\arabic{enumi}.}
\tightlist
\item
  \textbf{Instrumented collection is required.} The boundary exists
  because the agent was asked to declare it, so the method does not
  apply to logs recorded without the protocol, and we validate no
  procedure for inferring boundaries in them. The schema is also thin:
  we persist the next-declaration boundary and the auditor support
  calibration, not an explicit lifecycle end, so termination coverage is
  zero.
\item
  \textbf{The human annotation is an interested floor.} One non-expert
  annotator, 13 trajectories, 1.55x over-segmentation, with the mark
  budget and the primary statistic fixed after the fact and the bias
  running toward over-recovery. Every other label here is
  model-produced; only the lexical control and the mechanical rule
  depend on no model judgment.
\item
  \textbf{The boundary is not shown to be optimal for training.} Beating
  equal-length windows and ruling out a mechanical rule are attribution
  comparisons, not training ones; a downstream comparison changing only
  the boundary rule remains unrun.
\item
  \textbf{The construction attribution does not reach significance,}
  landing between \(p = 0.09\) and \(0.13\) over 60 items in 15
  instances, on sets that also differ in instance pool and difficulty.
  The training evidence covers one 27B family and one run per arm, 1,825
  of 2,551 pairs come from one resampling family under one generator,
  and the credit assigner is implemented but never trained.
\end{enumerate}

\section{Conclusion}\label{conclusion}

A declaration protocol buys a semantic boundary for one line of prompt,
where the alternatives need a gold patch, a milestone vocabulary, a
replayable environment, or a second model. That boundary is real: with
the declaration deleted, actions inside a phase stay attributable to the
hypothesis that governed them at over twice chance and better than
equal-length blocks over the same trajectories (paired \(p = 0.0002\)),
while a code-blind reader recovers positions only by over-segmenting and
a mechanical test-event rule does no better than chance --- and
expensive to reproduce afterwards, which is when declaring it during
collection earns its cost.

Its usefulness splits in two. As a construction interface it already
pays: four supervised targets from one collection, audit supervision
drawn from spans an episode label would flatten, and a credit assignment
differing in sign from the episode-level one on two thirds of
multi-phase successes. Whether a policy trained through it ends up
better is open: a standard preference optimizer fits the unit but learns
something bound to how the pairs were written, so the next step is a
more diverse pair corpus, not a different boundary.

\label{hl@bodyend}
\clearpage
\bibliography{references.bib}

\appendix
\section{Release and reproducibility}\label{release-and-reproducibility}

The project repository carries the materials this study was produced
from: phase extraction and evidence binding, phase audit and preference
construction, the SFT and DPO training configurations, the held-out
preference item sets with their construction code and per-item scoring
output for all four arms, the phase and pair schemas, the prompts and
masking rules, the figure-generation scripts, and the manifests,
instance IDs, hashes and regeneration commands. Source trajectories that
are not redistributable are represented by instance IDs, phase
annotations, derived pairs, exact schemas, artifact hashes, and the
commands that regenerate the private-to-derived transformation.

Code is at \url{https://github.com/Jingxi-Wei/hypothesis-ledger} and the
phase ledger with the derived training and evaluation views is at
\url{https://huggingface.co/datasets/jingxiwei/hypothesis-ledger-selfcorrection}.
The released ledger omits the trajectories collected under the
superseded unsanitized-feedback protocol, whose correction phases can
quote withheld test expectations: 761 of the 789 instances carrying both
a ledger and an audit survive that exclusion, and every released phase
record carries the holdout tag the training files were filtered against.
The code repository carries the pipeline and the paper's own measurement
scripts and artifacts, the latter under \texttt{paper/evidence/}; the
training and evaluation corpora are on the dataset rather than
duplicated in both places.

\begin{table}[!ht]
\textbf{Table A1: Canonical artifact map.} Paths are relative to the
code repository unless marked \emph{(dataset)}, which are files in the
Hugging Face dataset.

{\def\LTcaptype{none} 
\begin{center}
\begin{tabular}{@{}
>{\raggedright\arraybackslash}p{(\linewidth - 2\tabcolsep) * \real{0.5000}}
  >{\raggedright\arraybackslash}p{(\linewidth - 2\tabcolsep) * \real{0.5000}}@{}}
\toprule\noalign{}
\begin{minipage}[b]{\linewidth}\raggedright
Component
\end{minipage} & \begin{minipage}[b]{\linewidth}\raggedright
Artifact
\end{minipage} \\
\midrule
phase records & \texttt{ledger\_cards.jsonl} \emph{(dataset)}; sources
are \texttt{dataset/\allowbreak{}raw/\allowbreak{}*/\allowbreak{}*/\allowbreak{}\{trajectory,ledger,audit\}.json}, which are
not redistributable \\
collection protocol & \texttt{src/\allowbreak{}configs/\allowbreak{}swebench\_\allowbreak{}hypo.yaml} \\
extraction / SFT export & \texttt{src/export.py} \\
preference assembly and holdout wall &
\texttt{preference/prep\_rm.py} \\
SFT data & \texttt{sft\_train.jsonl} \emph{(dataset)} \\
Phase-DPO pairs / label-inverted / verified-only &
\texttt{rm\_\allowbreak{}pairs\_\allowbreak{}train\{,\_\allowbreak{}flip,\_\allowbreak{}verified\}.jsonl}
\emph{(dataset)} \\
DPO configurations &
\texttt{training/\allowbreak{}qwen27b\_\allowbreak{}dpo\{,\_\allowbreak{}flip,\_\allowbreak{}verified\}.yaml} \\
corpus census & \texttt{scripts/\allowbreak{}summarize\_\allowbreak{}phase\_\allowbreak{}dataset.py},
\texttt{artifacts/\allowbreak{}phase\_\allowbreak{}dataset\_\allowbreak{}statistics.json} \\
adversarial held-out items / scores &
\texttt{training/\allowbreak{}rm\_\allowbreak{}eval\_\allowbreak{}pairs\_\allowbreak{}heldout\_\allowbreak{}v2*.jsonl},
\texttt{training/\allowbreak{}dpo\_\allowbreak{}prefacc\_\allowbreak{}v2\_\allowbreak{}*.json} \\
matched-construction items / scores &
\texttt{training/\allowbreak{}rm\_\allowbreak{}eval\_\allowbreak{}pairs\_\allowbreak{}matched\_\allowbreak{}v3*.jsonl},
\texttt{training/\allowbreak{}dpo\_\allowbreak{}prefacc\_\allowbreak{}matched\_\allowbreak{}v3\_\allowbreak{}*.json} \\
attribution and paired-window probes &
\texttt{artifacts/\allowbreak{}phase\_\allowbreak{}assign\_\allowbreak{}api.jsonl},
\texttt{artifacts/\allowbreak{}phase\_\allowbreak{}coherence\_\allowbreak{}assignment.json} \\
mechanical test-event rules &
\texttt{artifacts/\allowbreak{}mechanical\_\allowbreak{}test\_\allowbreak{}event\_\allowbreak{}rule\{,\_\allowbreak{}strict\}.json} \\
rollout efficiency & \texttt{scripts/\allowbreak{}summarize\_\allowbreak{}rollout\_\allowbreak{}efficiency.py},
\texttt{artifacts/\allowbreak{}rollout\_\allowbreak{}efficiency\_\allowbreak{}three\_\allowbreak{}arm.json} \\
blinded phase audit &
\texttt{artifacts/\allowbreak{}phase\_\allowbreak{}model\_\allowbreak{}audit\_\allowbreak{}codex\_\allowbreak{}50.jsonl},
\texttt{...\_summary.json} \\
training diagnostics &
\texttt{scripts/\allowbreak{}summarize\_\allowbreak{}dpo\_\allowbreak{}training\_\allowbreak{}evidence.py},
\texttt{artifacts/\allowbreak{}dpo\_\allowbreak{}training\_\allowbreak{}diagnostics.json} \\
compression measurement & \texttt{scripts/\allowbreak{}measure\_\allowbreak{}compression.py},
\texttt{artifacts/\allowbreak{}compression.json} \\
declaration cost & \texttt{scripts/\allowbreak{}measure\_\allowbreak{}declaration\_\allowbreak{}cost.py},
\texttt{artifacts/\allowbreak{}declaration\_\allowbreak{}cost.json} \\
phase-level credit assignment & \texttt{phase\_\allowbreak{}rl/\allowbreak{}phase\_\allowbreak{}credit.py} \\
case studies & \texttt{manuscript/\allowbreak{}appendix\_\allowbreak{}case\_\allowbreak{}studies.md} \\
\bottomrule
\end{tabular}
\end{center}
}
\end{table}

Paths under \texttt{scripts/}, \texttt{artifacts/} and
\texttt{manuscript/} are relative to \texttt{paper/evidence/}; the rest
are relative to the repository root.

\section{Attribution controls in
full}\label{attribution-controls-in-full}

The headline sample is 221 phases from 49 trajectories with
\(3 \le K \le 12\), mean chance 0.222. Accuracy is roughly flat as
chance falls with \(K\): 0.485 at \(K{=}3\) against 0.333, and 0.542 at
\(K{=}12\) against 0.083. The model-free TF-IDF cosine reaches 0.353 on
the identical items.

The permutation and offset controls run on a second, larger sample of
502 phases from 51 trajectories that declare more hypotheses each, so
its chance rate is 0.102. Raw accuracies across the two samples are not
comparable and we quote multiples of chance. Unshuffled it reaches 0.219
(2.15x); shuffling which hypothesis belongs to which block collapses it
to 0.108 (1.06x).

\begin{table}[!ht]
\textbf{Table B1: Attribution accuracy when every boundary is slid by a
fixed offset.} Second sample, 502 phases, chance 0.102.

{\def\LTcaptype{none} 
\begin{center}
\begin{tabular}{@{}
>{\raggedright\arraybackslash}p{(\linewidth - 14\tabcolsep) * \real{0.0968}}
  >{\raggedleft\arraybackslash}p{(\linewidth - 14\tabcolsep) * \real{0.1290}}
  >{\raggedleft\arraybackslash}p{(\linewidth - 14\tabcolsep) * \real{0.1290}}
  >{\raggedleft\arraybackslash}p{(\linewidth - 14\tabcolsep) * \real{0.1290}}
  >{\raggedleft\arraybackslash}p{(\linewidth - 14\tabcolsep) * \real{0.1290}}
  >{\raggedleft\arraybackslash}p{(\linewidth - 14\tabcolsep) * \real{0.1290}}
  >{\raggedleft\arraybackslash}p{(\linewidth - 14\tabcolsep) * \real{0.1290}}
  >{\raggedleft\arraybackslash}p{(\linewidth - 14\tabcolsep) * \real{0.1290}}@{}}
\toprule\noalign{}
\begin{minipage}[b]{\linewidth}\raggedright
offset (messages)
\end{minipage} & \begin{minipage}[b]{\linewidth}\raggedleft
\(-8\)
\end{minipage} & \begin{minipage}[b]{\linewidth}\raggedleft
\(-4\)
\end{minipage} & \begin{minipage}[b]{\linewidth}\raggedleft
\(-2\)
\end{minipage} & \begin{minipage}[b]{\linewidth}\raggedleft
\(0\)
\end{minipage} & \begin{minipage}[b]{\linewidth}\raggedleft
\(+2\)
\end{minipage} & \begin{minipage}[b]{\linewidth}\raggedleft
\(+4\)
\end{minipage} & \begin{minipage}[b]{\linewidth}\raggedleft
\(+8\)
\end{minipage} \\
\midrule
accuracy & 0.165 & 0.223 & 0.271 & 0.219 & 0.169 & 0.153 & 0.125 \\
multiple of chance & 1.62 & 2.19 & 2.66 & 2.15 & 1.66 & 1.50 & 1.23 \\
\bottomrule
\end{tabular}
\end{center}
}
\end{table}

\begin{figure}
\centering
\includegraphics[width=0.78\linewidth,height=\textheight,keepaspectratio,alt={Attribution decays away from the declared position. Second control sample, 502 phases over 51 trajectories, chance 0.102. The maximum falls two messages before the declaration, which is what one expects when an agent recognizes a new direction and then writes it down.}]{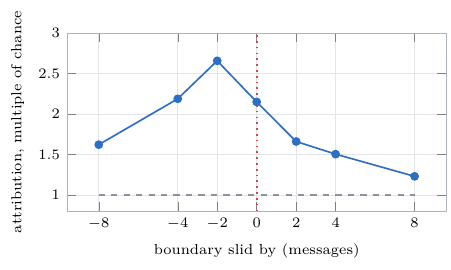}
\caption{Attribution decays away from the declared position. Second
control sample, 502 phases over 51 trajectories, chance 0.102. The
maximum falls two messages \emph{before} the declaration, which is what
one expects when an agent recognizes a new direction and then writes it
down.}\label{fig:offset}
\end{figure}

\begin{figure}
\centering
\includegraphics[width=1\linewidth,height=\textheight,keepaspectratio,alt={Paired difference by trajectory. Declared minus equal-length attribution accuracy for each trajectory, sorted. Both arms were scored by one matcher in one run, so any bias of the matcher cancels within a pair.}]{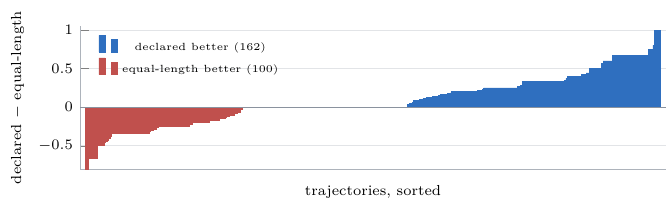}
\caption{Paired difference by trajectory. Declared minus equal-length
attribution accuracy for each trajectory, sorted. Both arms were scored
by one matcher in one run, so any bias of the matcher cancels within a
pair.}\label{fig:paired}
\end{figure}

\textbf{The paired window comparison.} Both arms were scored in one run
by one matcher over 374 trajectories: declared blocks 0.495 against
chance 0.208 (2.38x, 95\% trajectory-clustered {[}2.25, 2.51{]},
\(n = 1884\)); equal-length blocks 0.431 against 0.208 (2.07x, {[}1.96,
2.20{]}, \(n = 1917\)). The marginal intervals are disjoint: 162
declared-better, 100 window-better, 112 tied, sign test \(p = 0.0002\).
TF-IDF scores 0.316 and 0.305 on the two arms, so the lexical baseline
does not separate them; on declared blocks the model exceeds it by
+0.178 (95\% clustered {[}0.152, 0.206{]}).

\section{Held-out preference
statistics}\label{held-out-preference-statistics}

\begin{table}[!ht]
\textbf{Table C1: Per-arm results on the adversarial held-out set,} 91
items over 87 never-trained instances.

{\def\LTcaptype{none} 
\begin{center}
\begin{tabular}{@{}
>{\raggedright\arraybackslash}p{(\linewidth - 6\tabcolsep) * \real{0.2000}}
  >{\raggedleft\arraybackslash}p{(\linewidth - 6\tabcolsep) * \real{0.2667}}
  >{\raggedleft\arraybackslash}p{(\linewidth - 6\tabcolsep) * \real{0.2667}}
  >{\raggedleft\arraybackslash}p{(\linewidth - 6\tabcolsep) * \real{0.2667}}@{}}
\toprule\noalign{}
\begin{minipage}[b]{\linewidth}\raggedright
Arm
\end{minipage} & \begin{minipage}[b]{\linewidth}\raggedleft
Accuracy
\end{minipage} & \begin{minipage}[b]{\linewidth}\raggedleft
Decisions changed vs.~base
\end{minipage} & \begin{minipage}[b]{\linewidth}\raggedleft
Shift toward the advancing side
\end{minipage} \\
\midrule
SFT reference & 0.824 & --- & --- \\
Phase-DPO (main) & 0.824 & 0 / 91 & 0.505 \\
label-inverted & 0.824 & 0 / 91 & 0.440 \\
verified-only & 0.824 & 0 / 91 & 0.495 \\
\bottomrule
\end{tabular}
\end{center}
}
\end{table}

All four arms share the 95\% instance-clustered bootstrap interval
{[}0.744, 0.898{]} against a pick-the-longer floor of 0.626.

\textbf{Significance on the matched set.} Four of 60 decisions change
under the main arm and none under either control. Exact McNemar against
the arm's own baseline, \(b = 4\), \(c = 0\): \(p = 0.125\). Fisher
exact, 4/60 against 0/60: \(p = 0.119\). Instance-level sign test, 4
discordant instances of 15, all concordant in direction: \(p = 0.125\)
two-sided. Shift-direction rate \(37/60 = 0.617\): binomial
\(p = 0.093\). The last two ignore the clustering of 60 items into 15
instances, which can only weaken them. Median base margin is 18.89 on
the adversarial set and 14.56 on the matched set; the main arm's shift
is 1.40 median / 7.87 max and 4.05 / 40.62 respectively, while the
label-inverted arm's is 0.37 and 0.50.

\textbf{Optimization diagnostics.} Over the first-to-last 20\% of
training steps the main arm moves loss 0.6696 to 0.4510, reward margin
0.0550 to 1.2133, and train accuracy 0.619 to 0.798; the same-size
label-inverted arm ends at margin \(-0.0027\) and accuracy 0.473.

\begin{figure}
\centering
\includegraphics[width=0.86\linewidth,height=\textheight,keepaspectratio,alt={Why no decision changes on the adversarial set. Each point is one held-out item, placed by the adapter's shift as a fraction of the gap it would have to close; a decision can only change at or past 1. The adversarial items cluster an order of magnitude short of it.}]{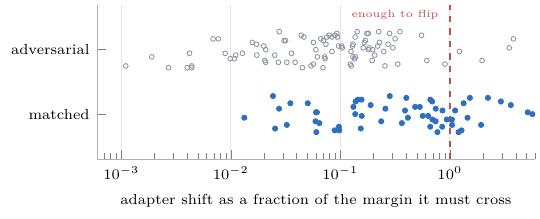}
\caption{Why no decision changes on the adversarial set. Each point is
one held-out item, placed by the adapter's shift as a fraction of the
gap it would have to close; a decision can only change at or past 1. The
adversarial items cluster an order of magnitude short of
it.}\label{fig:shiftratio}
\end{figure}

\section{The mechanical test-event
rule}\label{the-mechanical-test-event-rule}

The strict rule fires only on test-runner commands (\texttt{pytest},
\texttt{go\ test}, \texttt{npm\ test}, \texttt{tox}, and equivalents)
and places 41 marks over the 13 annotated trajectories, matching 4 of 40
declared boundaries against a random expectation of 5.6 for that many
marks, with no marks at all on 5 of the 13. The permissive rule fires on
any command the pipeline classifies as a check and places 273 marks,
matching 29 of 40 against a random expectation of 29.4. Matching is
greedy one-to-one within a \(\pm 1\) step tolerance, identical to the
human annotation scorer, and the random expectation is taken by
permuting the same number of marks over the same trajectory lengths.

\section{The phase-level credit
assigner}\label{the-phase-level-credit-assigner}

Each declared phase receives a verdict from the agent's own executed
checks: a non-zero return code or a failure signal in the output yields
FAIL, a clean zero return code with a pass signal yields PASS, and
output carrying neither is treated as a dump rather than a check and
yields no verdict. The verdict classifier is a deterministic parser over
shell output, not a model judgment, but we have not validated it against
the repository's real test suite; the 68\% figure above depends on it.

The reward ordering separates a surviving fix from one a later phase
overwrites, and a clean refutation from a span that reaches no verdict
at all. Its magnitudes are unvalidated design constants and no result in
this paper depends on them; only the ordering is claimed. Advantages are
group-normalized over the phases of one trajectory, which for an
already-collected trajectory stands in for the online form, where a
group would instead be \(G\) resampled continuations from the same
boundary state.

\section{Where local training units come
from}\label{where-local-training-units-come-from}

\begin{table}[!ht]
\textbf{Table F1: Where local training units come from.} ``Extra
dependency'' refers only to creating or validating the boundary, not to
all training compute. This is a structural comparison of published
methods, not a head-to-head quality result.

{\def\LTcaptype{none} 
\begin{center}
\begin{tabular}{@{}
>{\raggedright\arraybackslash}p{(\linewidth - 4\tabcolsep) * \real{0.3333}}
  >{\raggedright\arraybackslash}p{(\linewidth - 4\tabcolsep) * \real{0.3333}}
  >{\raggedright\arraybackslash}p{(\linewidth - 4\tabcolsep) * \real{0.3333}}@{}}
\toprule\noalign{}
\begin{minipage}[b]{\linewidth}\raggedright
Method
\end{minipage} & \begin{minipage}[b]{\linewidth}\raggedright
Boundary source
\end{minipage} & \begin{minipage}[b]{\linewidth}\raggedright
Extra dependency for locality
\end{minipage} \\
\midrule
P2T \citep{ma2026patches} & developer-patch-derived process graph & gold
developer patch, teacher guidance \\
BEACON \citep{wang2026beacon} & task-meaningful environment milestone &
environment-specific milestone indicator \\
GEAR \citep{li2026gear} & teacher--student divergence spike &
ground-truth-conditioned teacher, white-box logits \\
CompILE \citep{kipf2019compile} & latent segmenter fit to demonstrations
& a segmentation model trained on the domain \\
SDPO \citep{kong2025sdpo} & retrospectively selected dialogue segment &
post-hoc selection over a finished session \\
Agentic-DPO \citep{chen2026agenticdpo} & expert action contrasted at the
same state & expert actions, replayable states \\
\textbf{Hypothesis Ledger (ours)} & agent's declared causal-hypothesis
adoption & collection-time declaration protocol \\
\bottomrule
\end{tabular}
\end{center}
}
\end{table}

The distinction that matters for cost is the last column: every row
above ours needs something the trajectory does not contain -- a gold
patch, a milestone vocabulary, a replayable state, a teacher's logits,
or a model fit to the domain -- and can therefore only place a boundary
after collection has finished. The declaration is the only entry priced
at collection time.

\section{The data interface end to
end}\label{the-data-interface-end-to-end}

\begin{figure}
\centering
\includegraphics[width=0.92\linewidth,height=\textheight,keepaspectratio,alt={Hypothesis-defined phases as a data interface. A hypothesis lifecycle creates a variable-length semantic segment. Retrospective audit binds evidence and diagnoses the segment; masked pair construction converts the decision into a local preference unit.}]{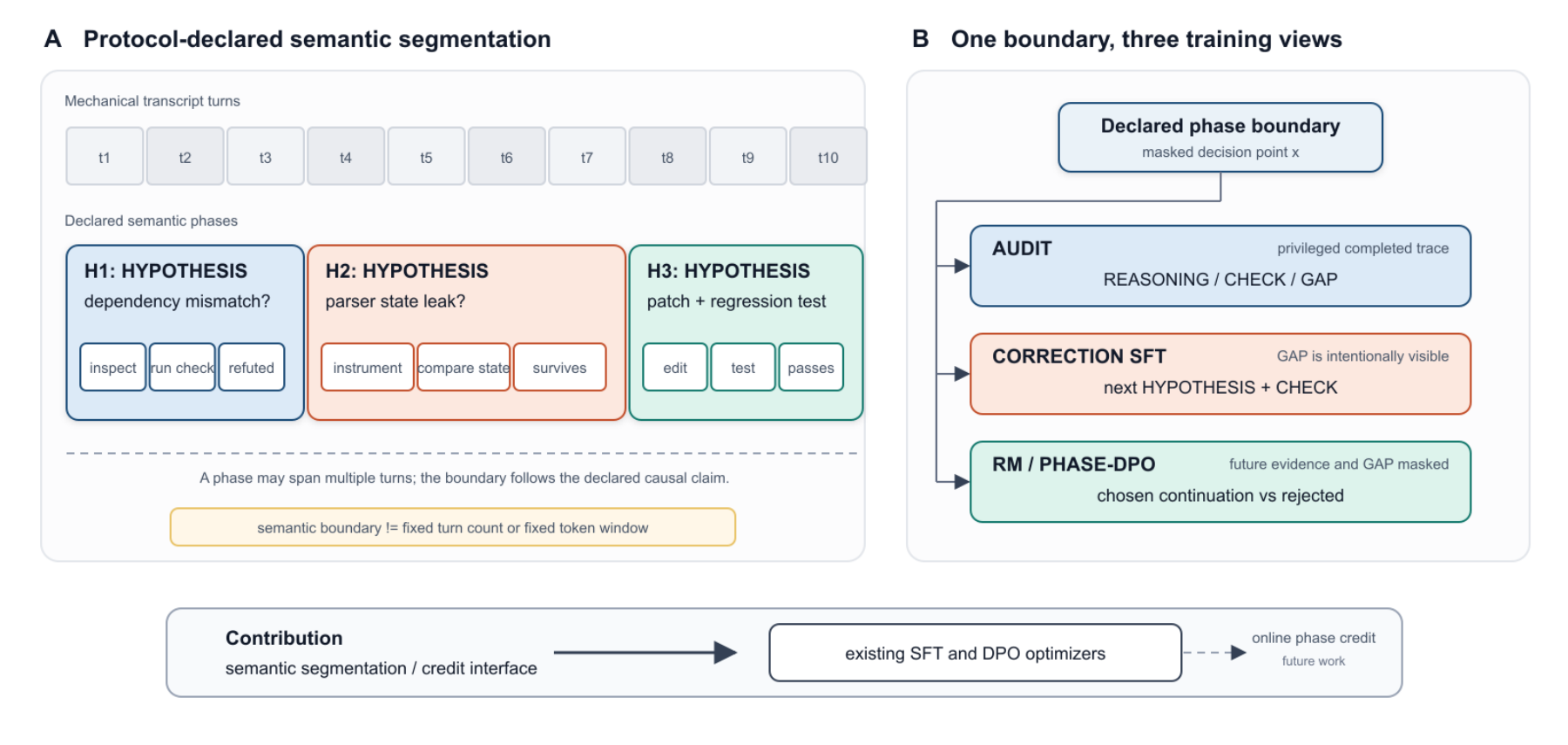}
\caption{Hypothesis-defined phases as a data interface. A hypothesis
lifecycle creates a variable-length semantic segment. Retrospective
audit binds evidence and diagnoses the segment; masked pair construction
converts the decision into a local preference unit.}\label{fig:pipeline}
\end{figure}

\section{One trajectory's phases}\label{one-trajectorys-phases}

\begin{table}[!ht]
\textbf{Table H1: Every phase of one trajectory,} from
\texttt{mozilla/bleach}, where an HTML sanitizer returned child nodes in
reverse order. ``Check'' is the verdict of the agent's own test inside
that phase, which is not the same as the hypothesis being right: H2
passes yet is superseded, and the cause is isolated only in H4.

{\def\LTcaptype{none} 
\begin{center}
\begin{tabular}{@{}
>{\raggedright\arraybackslash}p{(\linewidth - 6\tabcolsep) * \real{0.0562}}
  >{\raggedright\arraybackslash}p{(\linewidth - 6\tabcolsep) * \real{0.7865}}
  >{\raggedleft\arraybackslash}p{(\linewidth - 6\tabcolsep) * \real{0.0787}}
  >{\raggedleft\arraybackslash}p{(\linewidth - 6\tabcolsep) * \real{0.0787}}@{}}
\toprule\noalign{}
\begin{minipage}[b]{\linewidth}\raggedright
\end{minipage} & \begin{minipage}[b]{\linewidth}\raggedright
Declared hypothesis, abridged
\end{minipage} & \begin{minipage}[b]{\linewidth}\raggedleft
Span
\end{minipage} & \begin{minipage}[b]{\linewidth}\raggedleft
Check
\end{minipage} \\
\midrule
H1 & traversal consumes children last-in-first-out, reversing sibling
order & 18 & fails \\
H2 & \texttt{insertBefore} updates the tree but not
\texttt{\_childNodes} & 4 & passes \\
H3 & the vendored builder reports children differently from html5lib 1.1
& 4 & fails \\
H4 & \texttt{generateImpliedEndTags} pops the \emph{first} open element,
not the last & 23 & passes \\
\bottomrule
\end{tabular}
\end{center}
}
\end{table}

Nothing outside the declarations is needed to read this. Two of the four
directions are abandoned, the spans differ by a factor of six, and no
fixed window would separate them here.

\section{Extended related work}\label{extended-related-work}

Temporal abstraction represents extended behavior as options
\citep{sutton1999options}, and learned segmenters infer variable-length
subtasks from demonstrations \citep{kipf2019compile}. Agentic credit
assignment spans token, segment, step, turn, and multi-agent
granularities \citep{zhang2026credit}: a hierarchical credit module
\citep{luo2025agentlightning}, a planner/executor split scoring both
levels \citep{peng2026hiper}, and credit regions at student--teacher
divergence spikes \citep{li2026gear}. On the preference side a
retrospectively selected dialogue segment can serve as a unit between
turn and session \citep{kong2025sdpo}, and an expert action can be
contrasted with sampled negatives at the same state
\citep{chen2026agenticdpo}. What separates the present unit from all of
these is where the boundary comes from: each of them derives it from the
trajectory after the trajectory exists, whereas here the acting policy
emits it as part of the rollout.

\textbf{Tycho, and what converging with it does and does not settle.}
Tycho targets ARC-AGI-3, casting each game as a parameterized rendered
deterministic Moore machine and having a frontier coding agent
synthesize a per-game executable world model during play
\citep{lehmann2026tycho}. A hypothesis there is ``a free-form executable
program'' --- Python implementing \texttt{init\_state},
\texttt{transition}, \texttt{render} and \texttt{outcome} --- where ours
is a causal sentence checked by execution. It would be wrong to read
that as a structural difference. Both systems run one loop: the agent
commits to a conjecture, derives a prediction, receives the
environment's verdict, and either revises the conjecture or replaces it.
Their hypotheses are also agent-authored, recorded in notes before one
is chosen for encoding, so self-declaration is not what separates the
two either. What does is the audience: their declaration is an artefact
the agent then executes, ours is addressed to a protocol, which is what
lets an external reviewer negate one named conjecture and what leaves
the boundary recoverable afterwards. Two independent systems arriving at
the same organising structure from unrelated task morphologies is, we
think, evidence about the abstraction rather than about either
instantiation.

Their policy comparison also supplies external motivation for a choice
we made on other grounds. Their \texttt{Trigger} policy repairs the
model automatically whenever verification reports it unusable, and it
attains markedly better accepted transition match while playing worse
than actor-requested delegation --- about 83.07 against 88.49 mean
Relative Human Action Efficiency --- which they read as transition match
showing ``whether a simulator reproduces observed dynamics, not whether
it has identified the objective''. Patching a falsified conjecture along
its residual can therefore fit the observations better while moving away
from the cause. That is the band-aid failure mode our escalation ladder
is built to avoid: a refuted hypothesis is retired and the agent
redirected, not amended in place. The evidence is thinner than the
framing deserves and we do not lean on it: they report one matched run
per policy, their bootstrap intervals over the 25 games include zero for
the inter-policy contrasts, and those intervals test benchmark
composition rather than run-to-run variance. We read it as a single-run
observation consistent with our design, not as a settled result.

Where the two part company is consumption. Tycho is inference-time
orchestration and program synthesis over frontier models under matched
budgets; nothing is trained. Whether a hypothesis-scoped unit survives
being consumed as a post-training signal is untouched there and is the
question of our Section 6. Their decision set --- construct, repair,
query, plan through, bypass --- likewise leaves one question open that
their \texttt{Trigger} result raises: bypassing means declining to use
the model, not discarding the conjecture in favour of a successor, and
rival hypotheses are retained rather than eliminated from an explicit
version space. When a refuted hypothesis should be repaired and when it
should be abandoned is exactly the decision a phase boundary makes
observable, since abandonment becomes an event with a span and a verdict
attached to it.

\section{What a different contract would
change}\label{what-a-different-contract-would-change}

The extractor rewards neither brevity nor repair-complete spans, so a
contract asking for either would yield a different unit. Obtaining it
means changing the declaration contract and recollecting, not re-scoring
this corpus: the boundary is emitted during the rollout, so it cannot be
moved afterwards without moving what the agent did. Unit semantics are
configurable at collection time in that sense. We evaluate no
alternative contract here, and the comparison between contracts is the
experiment this design most obviously invites.

\end{document}